\pdfoutput=1
\documentclass[11pt]{article}
\usepackage[final]{acl}
\usepackage{times}
\usepackage{latexsym}
\usepackage[T1]{fontenc}
\usepackage[utf8]{inputenc}
\usepackage{microtype}
\usepackage{inconsolata}
\usepackage{graphicx}
\usepackage{kotex}
\usepackage{amsmath}
\usepackage{caption}

\usepackage{array}
\usepackage{float}
\usepackage{placeins}
\usepackage{multirow}

\usepackage{tabularx}
\usepackage{graphicx}

\usepackage{makecell}

\title{KITE: A Benchmark for Evaluating Korean Instruction-Following Abilities in Large Language Models}

\author{
  Dongjun Kim$^{1*}$ \\\And
  Chanhee Park$^{1}$ \\\And 
  Chanjun Park$^{2*}$ \\\And
  Heuiseok Lim$^{1}$ \\\AND
  $^{1}$Department of Computer Science and Engineering, Korea University\\$^{2}$School of Software, Soongsil University \\
  \texttt{\{junkim100, pch7678, limhseok\}@korea.ac.kr} \\
  \texttt{chanjun.park@ssu.ac.kr}\\
}

\begin{document}
\maketitle

\begin{abstract}
The instruction-following capabilities of large language models (LLMs) are pivotal for numerous applications, from conversational agents to complex reasoning systems. However, current evaluations predominantly focus on English models, neglecting the linguistic and cultural nuances of other languages. Specifically, Korean, with its distinct syntax, rich morphological features, honorific system, and dual numbering systems, lacks a dedicated benchmark for assessing open-ended instruction-following capabilities. To address this gap, we introduce the Korean Instruction-following Task Evaluation (KITE), a comprehensive benchmark designed to evaluate both general and Korean-specific instructions. Unlike existing Korean benchmarks that focus mainly on factual knowledge or multiple-choice testing, KITE directly targets diverse, open-ended instruction-following tasks. Our evaluation pipeline combines automated metrics with human assessments, revealing performance disparities across models and providing deeper insights into their strengths and weaknesses. By publicly releasing the KITE dataset\footnote{\href{https://huggingface.co/datasets/junkim100/KITE}{https://huggingface.co/datasets/junkim100/KITE}} and code\footnote{\href{https://github.com/junkim100/KITE}{https://github.com/junkim100/KITE}}, we aim to foster further research on culturally and linguistically inclusive LLM development and inspire similar endeavors for other underrepresented languages.
\end{abstract}

\section{Introduction}
Instruction-following (IF) abilities in Large Language Models (LLMs) are crucial, directly impacting the quality and effectiveness of applications ranging from conversational agents to complex decision-making systems~\cite{zhou2023instruction,kaddour2023challenges,liu2024llm}. IF capabilities enable LLMs to comprehend and execute tasks based on user instructions, making them integral to the performance of various natural language processing (NLP) applications. Despite the progress in developing robust IF capabilities, the evaluation of these abilities has predominantly focused on English-language models, resulting in a significant oversight of the linguistic and cultural nuances present in other languages~\cite{guo2023evaluating,chang2024survey}.

The current benchmarking efforts, although valuable, are heavily English-centric, which limits their applicability and relevance to non-English languages~\cite{wang2018glue,hendrycks2020measuring,srivastava2022beyond,wang2024mmlu}. This English-centric approach fails to capture the depth of linguistic and cultural diversity required for effective evaluation in multilingual contexts~\cite{costa2022no,wang2023challenging,weber2024investigating}. As a result, the performance assessment of LLMs in diverse linguistic environments remains inadequate, restricting their potential for broader applications in a multilingual world.

To address these limitations, we introduce Korean Instruction-following Task Evaluation (KITE), the first dedicated benchmark specifically designed to evaluate the Korean instruction-following capabilities of LLMs.

Korean presents unique challenges for LLMs due to its complex syntax, rich morphological features, and culturally embedded communication styles~\cite{lee2015introduction,park2021should,park2021klue}. Unlike English, Korean does not distinguish between uppercase and lowercase letters. Its agglutinative nature involves the combination of roots and affixes to form words and sentences, along with the use of post-positions, known as 'Josa,' which provides syntactic information about sentence components, allowing for flexible word order~\cite{park2014konlpy,park2020empirical,seo2023chef}. The honorific system and dual numbering systems (Sino-Korean and native Korean) add layers of complexity that necessitate dedicated evaluation benchmarks to accurately assess and improve the performance of LLMs in Korean instruction-following tasks.

By offering a detailed and culturally nuanced benchmark, this research provides a vital resource for advancing the development and evaluation of LLMs in Korean and inspiring similar efforts for other languages.

Our contributions are as follows:

\begin{itemize}
\item \textbf{Introduction of the KITE Benchmark}: We introduce KITE, a benchmark for evaluating Korean instruction-following in LLMs, addressing the gap in English-focused frameworks. The benchmark includes universally applicable and Korean-specific instructions, capturing the linguistic and cultural nuances of Korean, including syntax, honorifics, and dual number systems.

\item \textbf{Validation Through Diverse Experiments}: We validate KITE through a series of diverse experiments, providing valuable insights into the robustness and reliability of the benchmark. These experiments demonstrate the effectiveness of KITE in evaluating instruction-following capabilities across different NLP tasks and models.

\item \textbf{Public Release of KITE Benchmark}: We are making the KITE benchmark, including its code and data, publicly available. This resource aims to support researchers and practitioners in advancing the development and evaluation of culturally-aware LLMs for Korean, and to inspire the creation of similar benchmarks for other languages.
\end{itemize}

\begin{table*}[h]
    \centering
    \resizebox{\linewidth}{!}{
        \begin{tabularx}{\linewidth}{p{3cm}|X|X}
\hline
\multicolumn{1}{c|}{\textbf{Category}}                                                                              & \multicolumn{1}{c|}{\textbf{Description}}                                                                                             & \multicolumn{1}{c}{\textbf{Instructions}}                                                                                                                                                                              \\ \hline
\multirow{2}{*}{Acrostic Poem}                                                                                      & 모델이 주어진 단어의 각 글자로 시작하는 짧은 글을 생성하는 능력을 평가합니다.                                                                                          & "삼행시는 주어진 세 글자 단어의 각 글자로 시작하는 일관성 있는 짧막한 이야기야. '밤하늘'로 삼행시를 지어줘."                                                                                                                                                       \\
                                                                                                                    & \textit{Evaluates the model’s ability to generate structured poetry where each line starts with a specific letter from a given word.} & \textit{"An acrostic poem involves creating a short, coherent story where each line begins with a specific letter from a given three-character word. Please write an acrostic poem using the word '밤하늘' (night sky)."} \\ \hline
\multirow{2}{*}{Post-position Drop}                                                                                 & 모델의 한국어 문법 이해 능력을 평가하기 위해 조사 없이 문장을 형성하도록 요구합니다.                                                                                      & "주격 조사, 목적격 조사 없이 한글이 만들어진 기원에 대해서 설명하세요."                                                                                                                                                                             \\
                                                                                                                    & \textit{Tests the model’s understanding of Korean grammar by requiring sentences to be formed without postpositions.}                 & \textit{"Explain the origin of the Korean script without using subject or object postpositions."}                                                                                                                      \\ \hline
\multirow{2}{*}{Honorifics}                                                                                         & 모델이 존댓말과 반말 스타일 사이를 전환하는 능력을 평가합니다.                                                                                                   & "다음 문장을 반말로 바꿔 보세요: '어제 정말 즐거웠어요. 다음에 또 만나요.'"                                                                                                                                                                         \\
                                                                                                                    & \textit{Evaluates the model’s ability to switch between honorific and non-honorific styles of speech.}                                & "\textit{Convert the following sentence to informal speech: 'I had a really good time yesterday. Let's hang out again next time.'"}                                                                   \\ \hline
\multirow{2}{*}{\makecell[tl]{Native/Sino Korean\\ Number System}} & 모델이 순한국어와 한자어 숫자 체계 사이를 전환하는 능력을 평가합니다.                                                                                               & "다음 문장의 숫자를 순한국어로 바꾸세요. '이 회의는 90분 동안 지속됩니다.'"                                                                                                                                                                         \\
                                                                                                                    & \textit{Tests the model’s ability to switch between native and Sino-Korean number systems.}                                           & "\textit{Change the numbers in the following sentence to native Korean: 'This meeting lasts for ninety minutes.'"}                                                                                    \\ \hline
        \end{tabularx}
    }
    \caption{Overview of KITE Korean Benchmark. The benchmark consists of Categories, Descriptions, and Instructions.}
    \label{tab:KITE_Korean_overview}
\end{table*}

\section{Related Works}
Recent advancements in LLMs have shown significant progress in tasks that require instruction following~\cite{zhao2023survey}. Nevertheless, existing open benchmarks, such as IFEval~\cite{zhou2023instruction} and InstructEval~\cite{chia2023instructeval}, which have been crucial in evaluating instruction-following capabilities, are predominantly focused on English. This English-centric approach overlooks the nuances present in non-English languages, underscoring the necessity for benchmarks that accommodate linguistic diversity.

Language-specific benchmarks have been created to address the unique linguistic and cultural characteristics of different languages. For example, CLUE~\cite{xu2020clue} is a comprehensive benchmark for Chinese, while Kobbq~\cite{jin2023kobbq}, KorNAT~\cite{lee2024kornat}, KMMLU~\cite{son2024kmmlu}, CLIcK~\cite{kim2024click}, and HAE-RAE Bench~\cite{son2023hae} focus on Korean language models, incorporating cultural contexts. However, these benchmarks are not specifically designed to evaluate instruction-following capabilities.

The Korean language presents distinct challenges, such as complex syntax, an agglutinative structure, and the use of honorifics. These linguistic and cultural features necessitate a dedicated benchmark for accurate evaluation. Developing a benchmark specifically for Korean instruction-following tasks is essential to bridge this gap and ensure a more comprehensive assessment of LLMs capabilities in diverse linguistic contexts. This dedicated benchmark would not only improve the evaluation of existing models but also drive advancements in creating more culturally and linguistically aware LLMs, ultimately leading to better performance and applicability in non-English-speaking regions.

\section{KITE Benchmark}
\subsection{Overview of KITE Benchmark}
To comprehensively evaluate models in both general and Korean-specific contexts, we introduce the KITE benchmark, which is divided into two distinct versions: \textit{KITE General} and \textit{KITE Korean}. This dual approach allows us to assess models' performance across a wide range of tasks while addressing the unique challenges presented by the Korean language and culture.

The motivation for developing KITE lies in the need for a benchmark that goes beyond translating existing English datasets. While translated datasets can provide a baseline, they often fail to capture the intricate linguistic and cultural nuances of non-English languages. Therefore, KITE aims to provide a more representative and comprehensive evaluation framework for models in the Korean context.

KITE General consists of \textit{427 instructions}, focusing on universally applicable tasks derived from existing English IF datasets. In contrast, KITE Korean comprises \textit{100 instructions} created from scratch to address complex linguistic subtleties and cultural specificities unique to Korean. Recognizing the importance of verifiability in evaluation, we prioritized linguistic categories that allow for precise, rule-based assessment. Nevertheless, we have thoughtfully embedded Korean cultural context within these linguistic instructions to ensure a comprehensive and culturally relevant evaluation.

\subsection{Development of KITE General Benchmark}
\label{sec:KITE_general}
The development of the KITE General benchmark involves a two-step process: 1) automated translation and 2) contextual filtering. 

\paragraph{Translation Process}
The initial phase begins with the automated translation of an existing English instruction-following dataset using gpt-4o-2024-05-13, specifically the IFEval dataset proposed by \cite{zhou2023instruction}\footnote{\url{https://github.com/google-research/google-research/tree/master/instruction_following_eval/data}}. This translation is followed by a meticulous manual verification process to identify and correct any errors that could compromise the dataset's quality.

Maintaining alignment with established standards via translated benchmarks ensures consistent evaluation metrics and enables direct performance comparisons between English and Korean datasets. This consistency is crucial for assessing model robustness and adaptability across languages. Moreover, using translated benchmarks retains the rigorously tested instructions from the original dataset, offering a reliable baseline to identify specific strengths and challenges in Korean. 

\paragraph{Filtering Process}
We implemented a rigorous filtering process to ensure the translated instructions' relevance and accuracy. Given that the IFEval data was categorized, we systematically removed instructions from categories that are only relevant in English, such as those involving "capitalization". Five native Korean speakers, all graduate and undergraduate students with knowledge of NLP, participated in this filtering process, independently reviewing the instructions to determine their applicability within the Korean context. Details about the reviewers involved in this process can be found in Appendix~\ref{appendix:evaluators}

By requiring unanimous approval from all five members, we ensure  that all instructions included in KITE are contextually appropriate and linguistically relevant for evaluating instruction-following capabilities in Korean.

Out of the original 541 instructions, 114 were filtered out, resulting in a refined dataset of 427 instructions. Examples of filtered data can be found in Appendix \ref{appendix:appendix_filtered}.

\subsection{Development of KITE Korean Benchmark}
\paragraph{Linguistic and Cultural Analysis and Category Development}
This section outlines the systematic process undertaken to create our tailored dataset, ensuring its comprehensive linguistic and cultural relevance.

The foundational step in creating the KITE Korean benchmark involves a detailed analysis of the specific linguistic and cultural requirements of the Korean language. This process includes an in-depth examination of grammatical structures, linguistic features, cultural practices, and common instruction-following scenarios that realistically reflect Korean culture.

In collaboration with the authors and the five experts mentioned in section \ref{sec:KITE_general}, we identified and developed categories for the dataset, ensuring comprehensive coverage of these essential elements.

\paragraph{Korean Specialized Instructions}
To specifically address the unique grammatical and linguistic phenomena inherent to the Korean language, we incorporate four distinct instructions that are tailored to these features:

\begin{itemize}
\item \textbf{Acrostic Poem}: Analogous to the haiku generation instruction, but without the syllable count restriction, this instruction requires the model to generate poems with each line beginning with a specified syllable. This task evaluates the model's ability to produce coherent text under stringent constraints, reflecting a common literary practice in Korean.
\item \textbf{Post-position Drop}: This instruction is inspired by the frequent omission of Korean grammatical markers that indicate syntactic roles. It tests the model's ability to parse sentences, accurately identify parts of speech (POS), and generate grammatically correct sentences without the use of post-positions, while preserving the original meaning. This mirrors a common conversational nuance in Korean.
\item \textbf{Honorifics}: This instruction assesses the model's proficiency in switching between different levels of politeness, specifically between honorific and informal speech. By altering sentence endings to convey the same meaning across different levels of formality, the model's understanding of social hierarchy and politeness in Korean culture is evaluated.
\item \textbf{Native/Sino Korean Number System}: This instruction evaluates the model's capability to understand and interchangeably use the two number systems in Korean: native Korean numbers and Sino-Korean numbers. This reflects a critical aspect of Korean language usage, where different contexts dictate the use of different number systems.
\end{itemize}

For each of these four categories, we created 25 instructions, totaling 100 instructions for the KITE Korean benchmark. By integrating these specialized instructions, the KITE Korean benchmark effectively addresses the unique grammatical and linguistic features of the Korean language, which are absent in English. Table \ref{tab:KITE_Korean_overview} provides an overview of the KITE Korean benchmark. 

Given the difficulty of verifying cultural context through a rule-based evaluation, we focused on linguistic categories that are verifiable. Meanwhile, we carefully embedded the Korean cultural context within these linguistic instructions to ensure they are representative and relevant.

\section{Experiments}
\subsection{Experimental Design}
Our experiments are designed to rigorously evaluate the performance of LLMs in both general and Korean-specific instruction-following tasks, with a particular focus on the linguistic and cultural nuances unique to Korea. We tested a selection of generic and Korean-specific models across various shot settings, including zero-shot and few-shot (1 shot, 3 shot, 5 shot) scenarios. This approach allowed us to analyze the models' ability to generalize from limited examples, highlighting their strengths and weaknesses in different contexts.

To ensure robust evaluation, we conducted a human assessment to complement KITE automated scoring, validating the alignment between automated and human judgments. Additionally, we performed a correlation analysis with the Open Ko-LLM Leaderboard~\cite{park2024open} to contextualize KITE’s results within the broader landscape of established benchmarks. This analysis provided insights into the instruction-following capabilities of the models relative to other general NLP tasks. The human evaluation result can be found in Appendix \ref{appendix:humeval}.

\subsection{Model Details}
We evaluated the instruction-following capabilities of the following models: generic models such as gpt-3.5-turbo-0125 \cite{brown2020language}, gpt-4o-2024-05-13 \cite{achiam2023gpt}, Llama 3 8B Instruct~\footnote{\url{https://huggingface.co/meta-llama/Meta-Llama-3-8B-Instruct}}, and Gemma 7b Instruct \cite{team2024gemma}, as well as Korean-specific models including SOLAR 1 Mini Chat \cite{kim2023solar}, HyperCLOVA X 003 \cite{yoo2024hyperclova}, and EEVE v1.0 10.8b Instruct \cite{kim2024efficient}.

\subsection{Evaluation Details}
\label{sec:evaluation_details}
Similar to IFEval \cite{zhou2023instruction}, KITE employs \textit{verifiable instructions} to ensure clear and measurable outcomes for objective evaluation. KITE General instructions are derived from IFEval and are inherently verifiable. KITE Korean instructions are also designed to be verifiable, ensuring clarity and precision.

To assess model performance, we use a detailed accuracy scoring method. Each instruction \( I_i \) is decomposed into sub-instructions \( \{s_{i1}, s_{i2}, \ldots, s_{in_i}\} \). The model's response is evaluated for each sub-instruction \( s_{ij} \). If the response meets the criteria of the sub-instruction, it is marked as 'followed' (\( f(s_{ij}) = 1 \)). This rule-based checking ensures each part of a multi-part instruction is addressed correctly.

Overall accuracy is calculated as:
\begin{equation}
\text{Accuracy} = \frac{\sum_{i=1}^{N} \sum_{j=1}^{n_i} f(s_{ij})}{\sum_{i=1}^{N} n_i} \times 100\%
\end{equation}
where \( N \) is the total number of instructions, \( n_i \) is the number of sub-instructions for instruction \( i \), and \( f(s_{ij}) \) returns 1 if sub-instruction \( s_{ij} \) is followed correctly and 0 otherwise.

\begin{table*}[h]
    \centering
    \resizebox{\linewidth}{!}{
        \begin{tabular}{ccccccccccc}
        \hline
        \multicolumn{1}{c|}{\multirow{2}{*}{\textbf{Model}}} & \multicolumn{5}{c|}{\textbf{KITE General}}                                                                  & \multicolumn{5}{c|}{\textbf{KITE Korean}}                                             \\ \cline{2-11} 
        \multicolumn{1}{c|}{}                                & \textbf{0 Shot} & \textbf{1 Shot} & \textbf{3 Shot} & \textbf{5 Shot} & \multicolumn{1}{c|}{\textbf{Avg.}} & \textbf{0 Shot} & \textbf{1 Shot} & \textbf{3 Shot} & \textbf{5 Shot} & \textbf{Avg.} \\ \hline
        gpt-3.5-turbo-0125                                        & 75.92           & 82.87           & 85.18           & 83.79           & 81.91                              & 46.19           & 49.74           & 50.76           & 52.28           & 49.74         \\
        gpt-4o-2024-05-13                                               & \textbf{89.35}           & \textbf{90.74}           & \textbf{89.81}           & \textbf{91.20}           & \textbf{90.27}                              & \textbf{61.42}           & \textbf{64.97}           & \textbf{65.48}           & \textbf{65.98}           & \textbf{64.46}         \\
        Llama 3 8B IT                                        & 70.83           & 55.55           & 53.24           & 54.62           & 58.55                              & 51.77           & 40.60           & 45.68           & 45.68           & 45.93         \\
        Gemma 7b IT                                          & 76.85           & 56.94           & 53.70           & 58.33           & 61.45                              & 48.73           & 38.57           & 46.70           & 46.19           & 45.04         \\ \hline
        SOLAR 1 Mini Chat                                    & 46.29           & 60.18           & 64.81           & 67.12           & 59.60                              & 32.99           & 28.93           & 33.50           & 34.01           & 32.35         \\
        HyperCLOVA X 003                                     & 60.64           & 69.90           & 73.14           & 53.24           & 64.22                              & 45.68           & 47.71           & 50.76            & 48.22            &  48.09        \\
        EEVE v1.0 10.8b IT                                   & 75.92           & 56.48           & 53.70           & 54.62           & 60.18                              & 48.73           & 36.04           & 43.65           &  44.67               &  42.54             \\ \hline
        \end{tabular}
    }
    \caption{Scores for Different Task Categories: KITE General and KITE Korean}
    \label{tab:combined_scores}
\end{table*}

\subsection{Main Results}
Table \ref{tab:combined_scores} presents a comprehensive comparison of the performance of various models across KITE General and Korean benchmarks.

Several key observations can be drawn from these results. First, GPT-4o-2024-05-13 demonstrates high and consistent performance across both benchmarks, indicating strong generalization capabilities and robustness in handling diverse instructions. 

Secondly, despite being trained specifically for the Korean language, models such as SOLAR 1 Mini Chat, HyperCLOVA X 003, and EEVE v1.0 10.8b Instruct still fall short of the Korean language proficiency exhibited by GPT-4, indicating there is a considerable gap in achieving the language proficiency demonstrated by more advanced by models like GPT-4, and underscoring a critical need for further research and development in language-specific instruction following.

Thirdly, the observed performance disparities among the models highlight the necessity of incorporating diverse training data and employing sophisticated architectures to effectively handle the broad spectrum of instructions across both benchmarks. Our findings emphasize that, irrespective of strong performance in benchmarks such as MMLU~\cite{hendrycks2020measuring} or ARC~\cite{clark2018think}, achieving proficiency in instruction following mandates specialized tuning and targeted refinement. 

Lastly, our results indicate that the capabilities required for reasoning, commonsense knowledge, natural language understanding, hallucination prevention, and instruction following are distinct and must be addressed separately. This suggests that to excel in instruction following, it is crucial to perform dedicated tuning based on data specifically designed for this purpose.

\paragraph{Impact of Shot Settings on Instruction-Following Performance}
In our experiments, we randomly selected examples from the entire dataset to provide different shot settings. Contrary to the common expectation that performance improves with more shots, we found that in the IF task, the results were completely opposite. We speculate that this may be due to the varied nature of instructions among the examples given. 

Previous studies have not conducted detailed experiments based on different shot settings, making our study a novel contribution in this area. An important observation is the robust performance of GPT-4 across these shot variations. While most models exhibit fluctuating performance with changes in the number of shots, GPT-4 remains stable. This stability can be interpreted as a key indicator of its superior instruction-following capability.

Overall, our findings suggest that while shot settings can significantly impact model performance, models like GPT-4 that exhibit robustness across these settings are particularly valuable. This robustness in GPT-4’s performance is likely related to its advanced training and architectural design, preparing the model for diverse and dynamic instruction-following tasks. On the other hand, the fluctuating result from the Korean-specific LLMs, such as HyperCLOVA X 003 and EEVE v1.0 10.8b IT suggests that in-depth instruction tuning in various shot settings might be the key to the reliable performance of the models' instruction following abilities. The varied results across models highlight the need for further research into optimizing training methodologies and shot configurations to enhance performance in instruction-following tasks.

\section{Conclusion}
In this paper, we introduce KITE, a benchmark specifically designed to evaluate the open-ended instruction-following capabilities of Korean LLMs. KITE addresses the unique linguistic and cultural nuances of Korean, providing a comprehensive framework for assessing model performance on both general and Korean-specific instructions. Our evaluation highlights the strengths and weaknesses of various models and confirms KITE’s reliability through a high correlation between human and automated assessments. These findings underscore the importance of dedicated benchmarks like KITE in capturing the full spectrum of LLM capabilities in multilingual and cross-cultural contexts. We anticipate that KITE will drive further progress in Korean LLM development and ensure these models can be effectively used in real-world applications requiring instruction following. By sharing our benchmark and code, we aim to support additional cross-cultural NLP research and encourage the creation of similar benchmarks for other languages. Future work will involve expanding KITE to include a more diverse range of instruction types and additional languages. We also plan to incorporate real-world application scenarios to further validate model performance and utility in practical settings.

\section*{Limitations}
While KITE provides a comprehensive framework for evaluating the Korean instruction-following capabilities of LLMs, several limitations should be acknowledged:

\paragraph{Dataset Scope}
Although extensive, the KITE benchmark may not cover all linguistic and cultural nuances of the Korean language. The included instructions are representative but cannot encompass the entire spectrum of Korean linguistic and cultural contexts. Future work should consider expanding the dataset to include a broader range of scenarios and instructions to enhance its robustness and applicability.

\paragraph{Model Evaluation Diversity}
Our evaluation primarily focuses on a select number of LLMs, both generic and Korean-specific. While these models provide valuable insights, the diversity of models evaluated is limited. Assessing a wider range of models, including those fine-tuned for other tasks, could offer a more comprehensive understanding of instruction-following capabilities across different model architectures and training paradigms.

\paragraph{Human Evaluation Constraints}
The human evaluation was conducted by a limited number of experts and covered a subset of the full instruction set. While this provided valuable insights, a more extensive human evaluation involving a larger and more diverse pool of evaluators would be beneficial. This could help reduce potential biases and improve the reliability of the evaluation results.

\paragraph{Instruction Complexity}
KITE General and KITE Korean were designed to cover a range of instruction complexities. However, the complexity of instructions can vary significantly in real-world applications. Future iterations of KITE should include a more granular categorization of instruction complexity to better capture the varying levels of difficulty encountered in practical scenarios.

\paragraph{Generalizability to Other Languages}
While KITE provides a valuable framework for evaluating Korean instruction-following capabilities, the methodologies and insights derived may not be directly transferable to other languages with different linguistic structures and cultural contexts. Developing similar benchmarks tailored to the unique characteristics of other languages remains a critical area for future research.

\paragraph{Real-World Applicability}
The instructions used in KITE, although designed to be representative, may not fully reflect the diversity and complexity of real-world instructions encountered by LLMs in various applications. Future enhancements should focus on incorporating more realistic and application-specific instructions to better align the evaluation with practical use cases.

\section*{Ethics Statement}
In developing the KITE benchmark, we adhered to the highest ethical standards to ensure the integrity and fairness of our research. We ensured that the data collection and creation processes were conducted ethically, avoiding any inclusion of personal or sensitive information. The dataset was designed to be culturally sensitive and representative of the Korean language, minimizing the risk of propagating harmful stereotypes or biases.

All experiments were conducted under fair and controlled conditions. The selection of models and the experimental procedures were standardized to ensure unbiased evaluations. We maintained transparency and reproducibility by making the KITE framework, data, and code publicly available, allowing other researchers to replicate and build upon our work.

Our work is guided by a commitment to ethical AI development, aiming to address bias and fairness in AI systems. By providing a comprehensive and culturally sensitive evaluation framework, we strive to contribute positively to the advancement of fair and effective AI technologies. In conclusion, the KITE benchmark was developed with a strong commitment to ethical principles, ensuring our research upholds the highest standards of integrity, fairness, and transparency.

\bibliography{anthology}
\clearpage
\appendix
\onecolumn

\section{Details on Reviewers in the Data Creation and Evaluation Process}
\label{appendix:evaluators}
Table \ref{tab:evaluators} summarizes reviewer demographics, including birth year, gender, and major. Reviewers, born 1996 to 2000 with majors such as Computer Science and International Economics, brought strong, task-relevant backgrounds. Their expertise and rigorous review ensured the KITE Benchmark's quality and reliability.

\begin{table*}[h]
\centering
\scalebox{0.5}{
\resizebox{\linewidth}{!}{
\begin{tabular}{c|c|c|c}
\hline
\textbf{No.} & \textbf{Birth Year} & \textbf{Gender} & \textbf{Academic Major} \\ \hline
1            & 2000                & M               & Computer Science        \\ \hline
2            & 1996                & M               & Computer Science        \\ \hline
3            & 1999                & M               & Computer Science        \\ \hline
4            & 2000                & M               & International Economics \\ \hline
5            & 2000                & M               & International Economics \\ \hline
\end{tabular}
}
}
\caption{Demographic Information of Reviewers Involved in Data Creation and Evaluation}
\label{tab:evaluators}
\end{table*}

\section{Examples of Filtered English-Centric Instructions}
\label{appendix:appendix_filtered}

Table \ref{tab:filtered_examples} provides examples of English-centric instructions that were filtered out during the creation of the KITE Korean dataset.

\begin{table*}[h]
    \centering
    \resizebox{\linewidth}{!}{%
        \begin{tabularx}{\linewidth}{p{3.6cm}|>{\raggedright\arraybackslash}X}
\hline
\multicolumn{1}{c|}{\textbf{Category}} & \multicolumn{1}{c|}{\textbf{Instructions}} \\
\hline
\multirow{2}{*}{capital\_word\_frequency} 
  & "다음 중 물고기가 아닌 것은 무엇인가요: 연어 또는 아보카도? 답변에서 모든 단어를 대문자로 8단어 이하로 사용하세요." \par \textit{"Which of the following is not a fish: salmon or avocado? Please use no more than 8 words in all caps in your response."} \\
  & "난터킷 출신 남자에 대한 리머릭을 쓰세요, 표기법을 사용해 표현하고, 적어도 2개의 단어를 모두 대문자로 사용하세요." \par \textit{"Write a limerick about a guy from Nantucket, use notations to express it, and use at least 2 words with all capital letters."} \\
\hline
\multirow{2}{*}{english\_capital} 
  & "누군가의 건강을 위해 기도하는 것이 좋은 생각인가요? 당신의 대답은 모두 대문자로, 그리고 영어로 해야 합니다." \par \textit{"Is praying for someone's health a good idea? Your answer must be in all capital letters and in English."} \\
  & "영어로 그리고 모든 글자를 대문자로 Alvin and the Chipmunks에 대한 에세이를 쓰세요." \par \textit{"Write an essay about Alvin and the Chipmunks in English and in all capital letters."} \\
\hline
\multirow{2}{*}{english\_lowercase} 
  & "제 친구 바넷에 대해 모든 글자를 소문자로 사용하여 시를 써주세요." \par \textit{"Write a poem all in lowercase letters about my friend Barnet."} \\
  & "clifford blu 직원을 위한 스키마를 만드세요. 소문자만 사용하세요." \par \textit{"create a schema for a clifford blu employee. use only lowercase letters."} \\
\hline
\multirow{2}{*}{response\_language} 
  & "Pei Yao에게 화난 편지를 힌디어로만 써주세요. 다른 언어는 사용하지 마세요." \par \textit{"Write an angry letter to Pei Yao using only Hindi, no other language is allowed."} \\
  & "마라티어만 사용하여 출근급행에 관한 하이쿠 템플릿을 작성하십시오. 다른 언어는 허용되지 않습니다." \par \textit{"Write a haiku template about rushing to work using only the Marathi language, no other language is allowed."} \\
\hline
        \end{tabularx}%
    }
    \caption{Examples of Filtered English-Centric Instructions}
    \label{tab:filtered_examples}
\end{table*}

\begin{figure*}[t]
    \centering
    \resizebox{\linewidth}{!}{
    \includegraphics[width=1\linewidth]{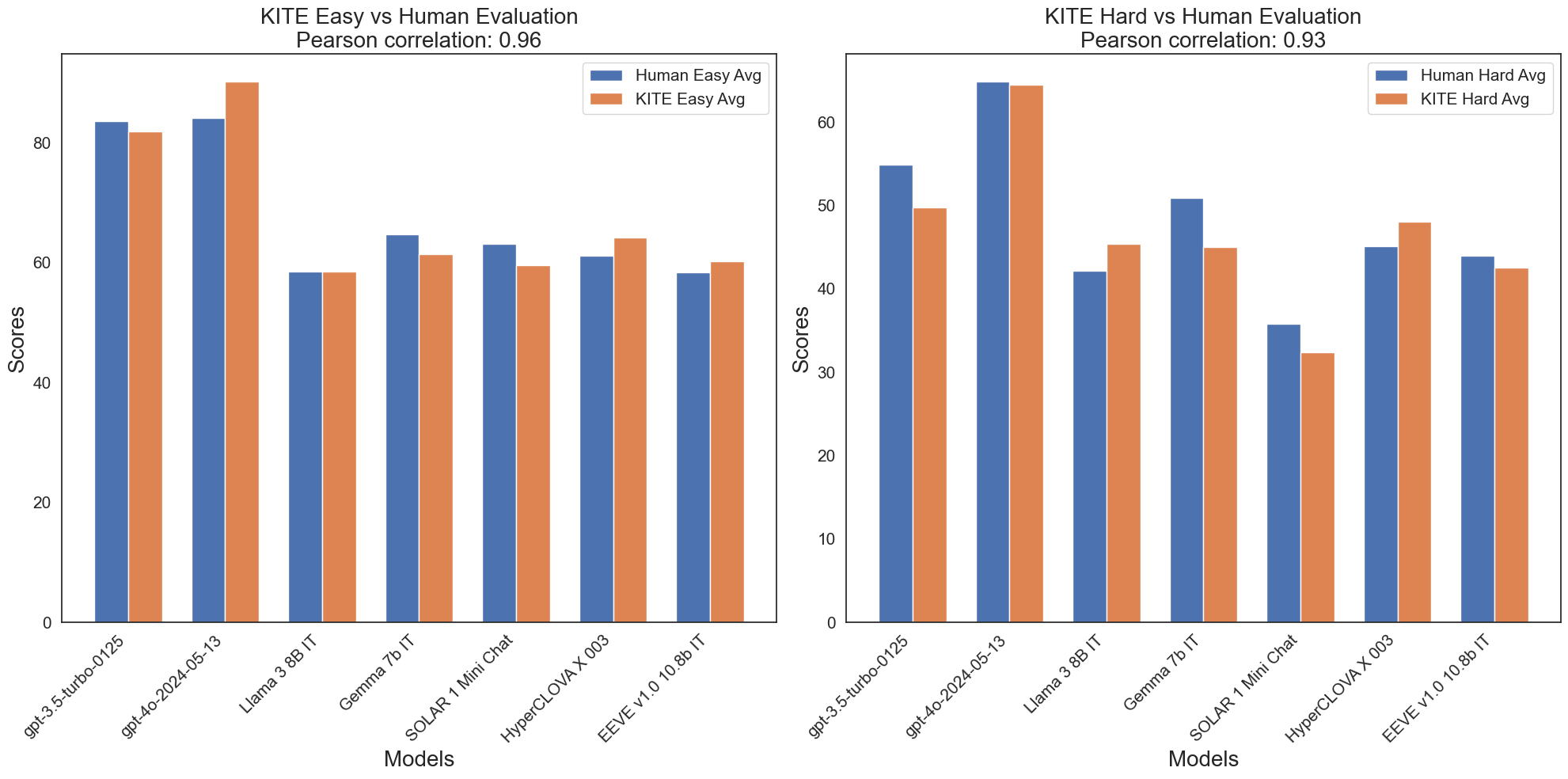}
    }
    \caption{Comparison of Human Evaluation and Automated KITE Scores (automated evaluation scores by the KITE Benchmarks). The left bar graph depicts the scores for KITE General, while the right bar graph illustrates the scores for KITE Korean. Each pair of bars represents the average scores assigned by human evaluators and the automated KITE Benchmarks for each model, demonstrating a strong agreement between the two evaluation methods.}
    \label{fig:human_vs_automated}
\end{figure*}

\section{Details on Human Evaluations}
\label{appendix:humeval}

\paragraph{Evaluation Setting} 
To compare the shot capabilities of models and humans, we uniquely provided humans with shot examples and conducted evaluations accordingly. We conducted a study involving five expert reviewers (mentioned in section \ref{sec:KITE_general}), who evaluated a subset of the instructions. Due to limited availability, 20\% of the full instruction set was randomly selected for manual evaluation in each section. Similar to the automated evaluation conducted by the KITE Benchmark, the human evaluators calculated the percentage of sub-instructions correctly followed by the models.

\paragraph{Evaluation Results}
Table~\ref{tab:human_evaluation} presents the average scores from five human evaluators for the response generated under each shot setting. Similar to the previous experiment, GPT-4o-2024-05-13 shows a consistent performance across different shot settings both in General and Korean benchmarks, whereas Korean-specific models struggle with the KITE Korean benchmark, demonstrating the difficulty of our benchmark.

Moreover, human evaluations show no significant change in performance across different shot settings. This consistency, despite varying example numbers, highlights a fundamental difference in the reasoning process between humans and models. It suggests that instruction-following tasks present significant challenges for current models, indicating a need for improvement in leveraging few-shot learning for better instruction-following.

\begin{table*}[h!]
    \centering
    \resizebox{\linewidth}{!}{
        \begin{tabular}{ccccccccccc}
        \hline
        \multicolumn{1}{c|}{\multirow{2}{*}{\textbf{Model}}} & \multicolumn{5}{c|}{\textbf{KITE General}}                                                                  & \multicolumn{5}{c|}{\textbf{KITE Korean}}                                             \\ \cline{2-11} 
        \multicolumn{1}{c|}{}                                & \textbf{0 Shot} & \textbf{1 Shot} & \textbf{3 Shot} & \textbf{5 Shot} & \multicolumn{1}{c|}{\textbf{Avg.}} & \textbf{0 Shot} & \textbf{1 Shot} & \textbf{3 Shot} & \textbf{5 Shot} & \textbf{Avg.} \\ \hline
        gpt-3.5-turbo-0125                                   & 83.00            & 83.50            & \textbf{84.20}            & 83.70            & 83.60                             & 54.50            & 55.00            & 54.80            & 55.20            & 54.88         \\
        gpt-4o-2024-05-13                                    & \textbf{84.00}            & \textbf{84.30}            & 84.10            & \textbf{84.20}            & \textbf{84.15}                             & \textbf{64.70}            & \textbf{65.00}            & \textbf{65.10}            & \textbf{64.80}            & \textbf{64.90}         \\
        Llama 3 8B IT                                        & 58.00            & 58.50            & 59.00            & 58.70            & 58.55                             & 42.00            & 42.30            & 42.20            & 42.10            & 42.15         \\
        Gemma 7b IT                                          & 64.50            & 64.80            & 64.90            & 64.60            & 64.70                             & 50.70            & 51.00            & 51.10            & 50.90            & 50.93         \\ \hline
        SOLAR 1 Mini Chat                                    & 63.00            & 63.20            & 63.30            & 63.10            & 63.15                             & 35.50            & 36.00            & 36.10            & 35.80            & 35.85         \\
        HyperCLOVA X 003                                     & 61.00            & 61.20            & 61.30            & 61.10            & 61.15                             & 45.00            & 45.20            & 45.30            & 45.10            & 45.15         \\
        EEVE v1.0 10.8b IT                                   & 58.00            & 58.50            & 58.90            & 58.40            & 58.45                             & 43.70            & 44.00            & 44.10            & 43.90            & 43.93         \\ \hline
        \end{tabular}
    }
    \caption{Human Evaluation Scores for Different Task Categories. The scores represent the average of all evaluators.}
    \label{tab:human_evaluation}
\end{table*}

\begin{figure}[t]
    \centering
    \includegraphics[width=0.7\linewidth]{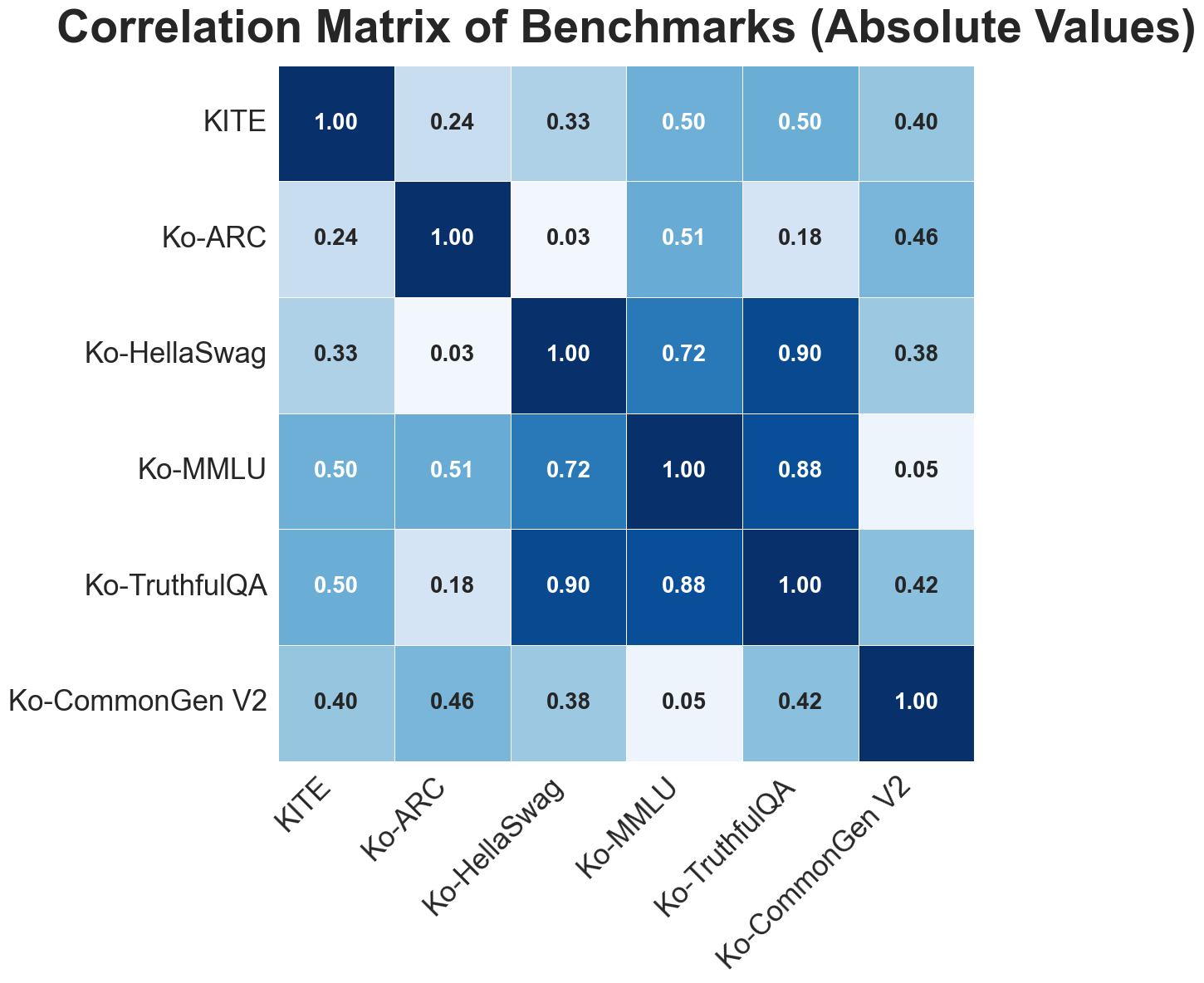}
    \caption{Correlation Matrix Between KITE and Other Common Korean Benchmarks}
    \label{fig:correlation_matrix}
\end{figure}

\paragraph{Correlation with Human and Automatic Evaluation}
To compare the human evaluation scores with the automated evaluation scores provided by the KITE Benchmarks, we conducted a statistical analysis using the Pearson correlation coefficient~\cite{benesty2009pearson}. Figure~\ref{fig:human_vs_automated} illustrates the comparison between human evaluation scores and KITE automated evaluation scores across both the General and Korean categories.

The Pearson correlation coefficients are \( r = 0.96 \) for KITE General and \( r = 0.93 \) for KITE Korean, indicating a strong agreement between the two evaluation methods. This high correlation indicates that the KITE Benchmarks effectively and reliably assess instruction-following capabilities in Korean, comparable to human evaluation.

\paragraph{Correlation with Open Ko-LLM Leaderboard}
We evaluated the top 10 instruction-tuned models from the Open Ko-LLM Leaderboard~\cite{park2024open}, which includes benchmarks such as Ko-Arc, Ko-Hellaswag, Ko-MMLU, Ko-TruthfulQA, and Ko-Commongen v2. These benchmarks, however, are not specifically designed to measure instruction-following capabilities.

To understand the relationship between instruction-following performance and other benchmark scores, we calculated the correlation between KITE scores (automated evaluation scores from the KITE Benchmarks) and the scores from the other benchmarks used in the Open Ko-LLM Leaderboard using the Pearson correlation coefficient~\cite{benesty2009pearson}.

Figure~\ref{fig:correlation_matrix} presents the correlation matrix, displaying the Pearson correlation coefficients between KITE and other common Korean benchmarks. The matrix values indicate the strength of the relationship between the scores from different benchmarks.

Key observations include a moderate positive correlation (\( r = 0.50 \))  between KITE and Ko-MMLU, suggesting that models excelling in instruction-following tasks also perform well across a range of tasks. Similarly, there is a moderate positive correlation (\( r = 0.50 \)) between KITE and Ko-TruthfulQA, implying that instruction-following capabilities may enhance truthful and accurate responses. Conversely, the correlation between KITE and Ko-HellaSwag is lower (\( r = 0.33 \)), indicating that KITE measures skills distinct from those required for commonsense reasoning. The low correlation with Ko-ARC (\( r = 0.24 \)) urther emphasizes that instruction-following and question-answering capabilities are not proportional. The moderate correlation with Ko-CommonGen V2  (\( r = 0.40 \)) suggests some relationship between instruction-following and coherent sentence generation from given concepts.

These varying degrees of correlation highlight the unique aspects of instruction-following capabilities. While moderate correlations exist with some benchmarks, lower correlations with others indicate that instruction-following captures different model behaviors not fully assessed by traditional benchmarks. These findings underscore the importance of dedicated benchmarks like KITE to evaluate instruction-following capabilities, providing additional insights into model performance that complement existing benchmarks.

\paragraph{Comparison of Model Performance on KITE and Human Evaluation}
\label{appendix:model_graph}

\begin{figure*}[h]
    \centering
    \resizebox{0.75\linewidth}{!}{
    \includegraphics[width=1\linewidth]{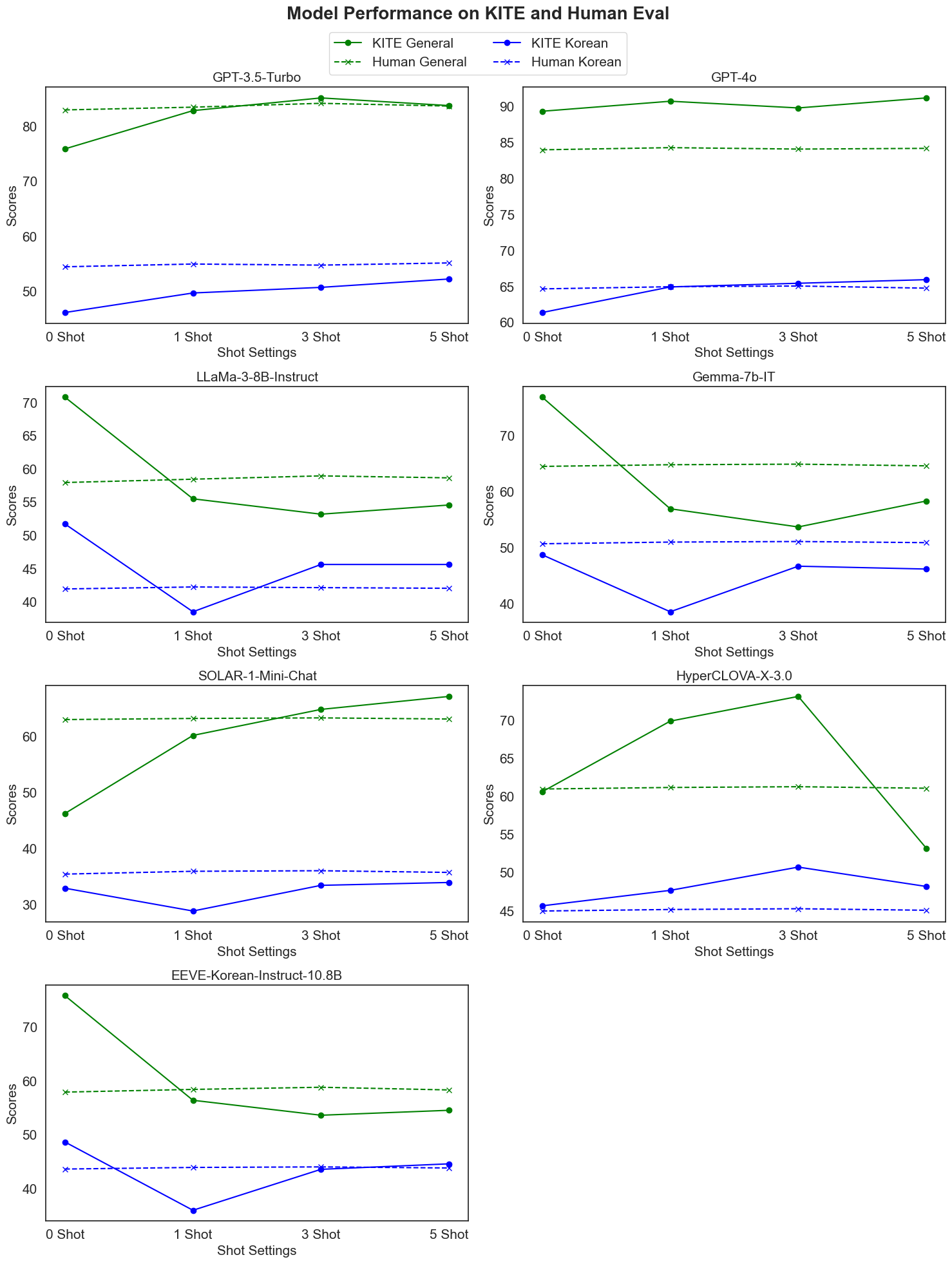}
    }
    \caption{Comparison of Model Performance on KITE and Human Evaluations across Different Shot Settings.}
    \label{fig:model_graph}
\end{figure*}

Figure \ref{fig:model_graph} presents a comprehensive comparison of model performance across different shot settings for both KITE and Human evaluations. Each subplot represents a different model, showcasing scores for both KITE General and KITE Korean evaluations against Human General and Human Korean evaluations. The solid lines indicate the KITE evaluation scores, with green lines representing General and blue lines representing Korean. Similarly, the dashed lines depict the Human evaluation scores, with green lines for General and blue lines for Korean. This visualization highlights the consistency and variance in model performance across the different shot settings and datasets, providing a clear comparison of each model's capability in handling instruction-following tasks in both general and Korean-specific contexts.

\end{document}